\ifdefined\XeTeXrevision\else\pdfoutput=1\fi
\documentclass[10pt,letterpaper]{article}
\usepackage[margin=0.92in,top=0.92in,bottom=0.88in,headheight=19pt]{geometry}
\usepackage[T1]{fontenc}
\usepackage[utf8]{inputenc}
\usepackage{lmodern}
\usepackage{microtype}
\usepackage{graphicx}
\usepackage{booktabs}
\usepackage{amsmath,amssymb}
\usepackage{xcolor}
\usepackage{fancyhdr}
\usepackage[font=small,labelfont=bf,labelsep=period]{caption}
\usepackage[colorlinks=true,linkcolor=inkblue,citecolor=inkblue,urlcolor=inkblue]{hyperref}

\definecolor{inkblue}{HTML}{1D4E63}
\definecolor{muted}{HTML}{4B5563}
\definecolor{rulegray}{HTML}{D1D5DB}

\hypersetup{
  pdftitle={OPUS: Selective reuse in continual learning of scientific field operators},
  pdfauthor={Samrendra Roy, Yoon Pyo Lee, Kazuma Kobayashi, Jay Phil Yoo, Souvik Chakraborty, Syed Bahauddin Alam},
  pdfsubject={cs.LG; physics.comp-ph; physics.flu-dyn}
}

\renewcommand{\headrulewidth}{0.35pt}
\renewcommand{\headrule}{\hbox to\headwidth{\color{rulegray}\leaders\hrule height \headrulewidth\hfill}}
\fancypagestyle{plain}{
  \fancyhf{}
  \renewcommand{\headrulewidth}{0pt}
  \fancyfoot[C]{\footnotesize\color{muted}\thepage}
}

\title{\vspace{-1.2em}\bfseries OPUS: Selective reuse in continual learning of\\ scientific field operators}
\author{%
  Samrendra Roy$^{1}$, Yoon Pyo Lee$^{1}$, Kazuma Kobayashi$^{1}$, Jay Phil Yoo$^{1}$,\\
  Souvik Chakraborty$^{1,2,3}$, and Syed Bahauddin Alam$^{1,4,\dagger}$\\[6pt]
  \small $^{1}$Department of Nuclear, Plasma, and Radiological Engineering,\\[-1pt]
  \small University of Illinois Urbana-Champaign, Urbana, IL 61801, USA\\
  \small $^{2}$Department of Applied Mechanics, Indian Institute of Technology Delhi,\\[-1pt]
  \small Hauz Khas, New Delhi 110016, India\\
  \small $^{3}$Yardi School of Artificial Intelligence (ScAI), Indian Institute of Technology Delhi,\\[-1pt]
  \small Hauz Khas, New Delhi 110016, India\\
  \small $^{4}$National Center for Supercomputing Applications (NCSA),\\[-1pt]
  \small University of Illinois Urbana-Champaign, Urbana, IL 61801, USA\\
  \small $^{\dagger}$Corresponding author: \href{mailto:alams@illinois.edu}{\texttt{alams@illinois.edu}}
}
\date{November 2026}

\begin{document}
\maketitle
\thispagestyle{plain}

\begin{abstract}
We introduce \textbf{OPUS} (\textbf{O}perator \textbf{P}rimitives for \textbf{U}nified \textbf{S}imulation), a modular unrolled operator library trained sequentially on heterogeneous field pairs. Using source-removal interventions, in which we disable one prior source at a time and refit on identical support fields without target physics labels, we test whether few-example adaptation causally depends on a specific prior source in a sequentially grown seven-task library (\textbf{V2}). Before permanent enrollment of Wave-Gauss, three labeled examples achieve 16.63\% mean relative \(L_2\) error across 32 paired draws versus 36.49\% for scratch training. Across 64 paired draws on the same pre-enrollment checkpoint, the intact library scores 16.80\%, disabling the earlier Wave-Layer source raises error by 28.47 percentage points (pp) to 45.27\%, and disabling Darcy leaves error nearly unchanged at 16.74\% (\(-0.06\)~pp). On Acoustic Mazes (24 draws), disabling Acoustic Inclusions raises error from 65.73\% to 98.24\% (\(+32.50\)~pp), whereas disabling Wave-Layer gives 64.38\% (\(-1.36\)~pp). To map empirical boundaries beyond \(32\times32\) grids, we evaluate two separate higher-resolution architectures (\textbf{V3} and \textbf{V4}) across multi-physics simulations, global reanalysis, ocean anomalies and \(\sim\)180k-node unstructured airfoil meshes. At full data, continual and scratch models perform comparably (scratch is marginally lower on four of seven V2 tasks); stored-route replay is exact by construction, whereas re-identification from fresh support can drift; and on unstructured airfoils, frozen learned blocks beat scrambled blocks on velocity at 200 training cases (7.30\% versus 8.05\%) yet trail a scratch GINO (7.01\%) and rank last of four arms at 25 cases. These findings provide controlled evidence of source-specific reuse on related few-example tasks while delineating where frozen reuse trails scratch training.
\end{abstract}

\noindent{\small\textbf{Keywords:} continual learning, neural operators, scientific machine learning, source-removal intervention, selective transfer, multi-geometry physics, unstructured meshes.}

\section{Introduction and related work}
\label{sec:intro}

Scientific archives span heterogeneous field counts, physical units, domain geometries and temporal horizons. Neural operators such as DeepONet \cite{deeponet}, the Fourier neural operator (FNO) \cite{fno}, the wavelet neural operator (WNO) \cite{wno} and geometry-informed neural operators (GINO) \cite{gino} have enabled surrogate modeling, virtual sensing and real-time digital twins across engineering and energy systems \cite{kobayashi2024,kobayashi2025rad,kobayashi2025conf,yoo2025sensing,yoo2026energy,roy2026active,roy2026adv,roy2026wave,verma2026flowmesh}. Recent scientific foundation models such as MPP \cite{mpp}, Poseidon \cite{poseidon}, DPOT \cite{dpot}, CoDA-NO \cite{codano}, Walrus \cite{walrus}, Aurora \cite{aurora} and SCNO \cite{roy2026scno} demonstrate multi-task pretraining across diverse PDE benchmarks such as PDEBench \cite{pdebench}, PDEArena \cite{pdearena}, APEBench \cite{apebench}, \emph{The Well} \cite{well} and global Earth reanalysis \cite{era5,wb}, with follow-up work stabilizing their autoregressive rollouts \cite{park2026healing} and embedding them in domain-specific agentic frameworks \cite{lee2025agentic}. Yet joint pretraining mixes all source datasets inside shared weights: a favorable comparison against training from scratch leaves unresolved whether downstream gains come from generic spatial smoothing, initialization, input preprocessing or reuse of a specific learned physical response, motivating diagnostic audits beyond standard training conventions \cite{lee2025feedback} and controlled transfer studies \cite{yoo2026gs}.

Continual and modular learning offers a way to separate storage from reuse. Regularization and pruning methods such as elastic weight consolidation (EWC) \cite{ewc}, PackNet \cite{packnet} and inter-spike-interval regularization \cite{roy2026isi} protect earlier tasks within fixed capacity. Modular and compositional approaches such as Progressive Neural Networks \cite{progressive}, compositional lifelong learning \cite{mendez2021}, task-driven modular networks \cite{veniat2021}, local module composition \cite{ostapenko2021}, the neural combinatorial wavelet neural operator (NCWNO) \cite{ncwno} and neuromorphic continual learners \cite{roy2026neurocl} instead differ in whether they freeze, update or append modules when learning a new task. Progressive Neural Networks perturb individual source columns to estimate transfer dependence \cite{progressive}, neural operator splitting composes frozen operator dictionaries at test time \cite{splitting}, and our companion GEODE study audits parameter isolation, route substitution and norm-matched randomized libraries across heterogeneous engineering interfaces \cite{geode}.

Here we ask a narrower, directly testable question for continual field-to-field operators: \emph{when a library grows stage by stage, does few-example adaptation to a new physical target causally depend on a specific prior source?} In OPUS, every frozen input stem and operator block retains its acquisition-stage identity. On a new target supplied only as three labeled field pairs with no target physics label (no task identifier or PDE label), we adapt routing and readout parameters, then disable one prior source at a time and repeat the exact adaptation on identical support and query fields. The defining signature is \emph{asymmetry}: removing a physically related source degrades accuracy to near random-library levels, whereas removing the designated comparison source changes mean error little.

\paragraph{Terminology.}
Throughout the paper we distinguish four operational concepts: (i) \emph{enrollment} adds a new stage's input stem and trainable operator blocks to the library after freezing all prior stages; (ii) \emph{stored task settings} record the normalization statistics, eligible block indices \(J_s\), mixing weights and readout fitted at enrollment; (iii) \emph{canonical replay} evaluates an earlier task using its stored task settings (exact preservation by construction); and (iv) \emph{fresh support} re-identifies routing and readout weights from a small newly sampled set of labeled field pairs without accessing stored task settings.

\paragraph{Scope of the three implementations.}
To separate our primary causal intervention from scaling and geometry boundary tests, we evaluate checkpoints from three distinct implementations. \textbf{V2} (Sections~\ref{sec:arch}--\ref{sec:full}) provides our main mechanistic result: paired source-removal interventions, full-data permanent enrollment across seven tasks on a shared \(32\times32\) grid, fresh-support retention, regime identification and withheld-parameter composition. \textbf{V3} (Sections~\ref{sec:unseen} and~\ref{sec:airfoil}) and \textbf{V4} (Section~\ref{sec:v4}) are boundary-mapping experiments with separate training histories: V3 tests a 14.77M-parameter library on enrolled NOAA ocean anomalies, unseen families from \emph{The Well} and \(\sim\)180k-node unstructured AirfRANS meshes, while V4 couples fine- and coarse-grid operator updates across compressible shocks, \(4:1\) heated-channel convection and global atmospheric fields on an equiangular latitude--longitude grid. Inspection of 28 V3 blocks showed that the spectral input projections had mean effective rank (exponentiated singular-value entropy) 15.65 out of 16 available directions, motivating V4's expanded dual-resolution design; because multiple architectural choices changed together between V3 and V4, we report V3 and V4 as descriptive boundary maps rather than isolated ablations.

\section{Architecture, continual library growth and methods}
\label{sec:arch}

\begin{figure}[!t]
  \centering
  \includegraphics[width=\linewidth,height=5.25in,keepaspectratio]{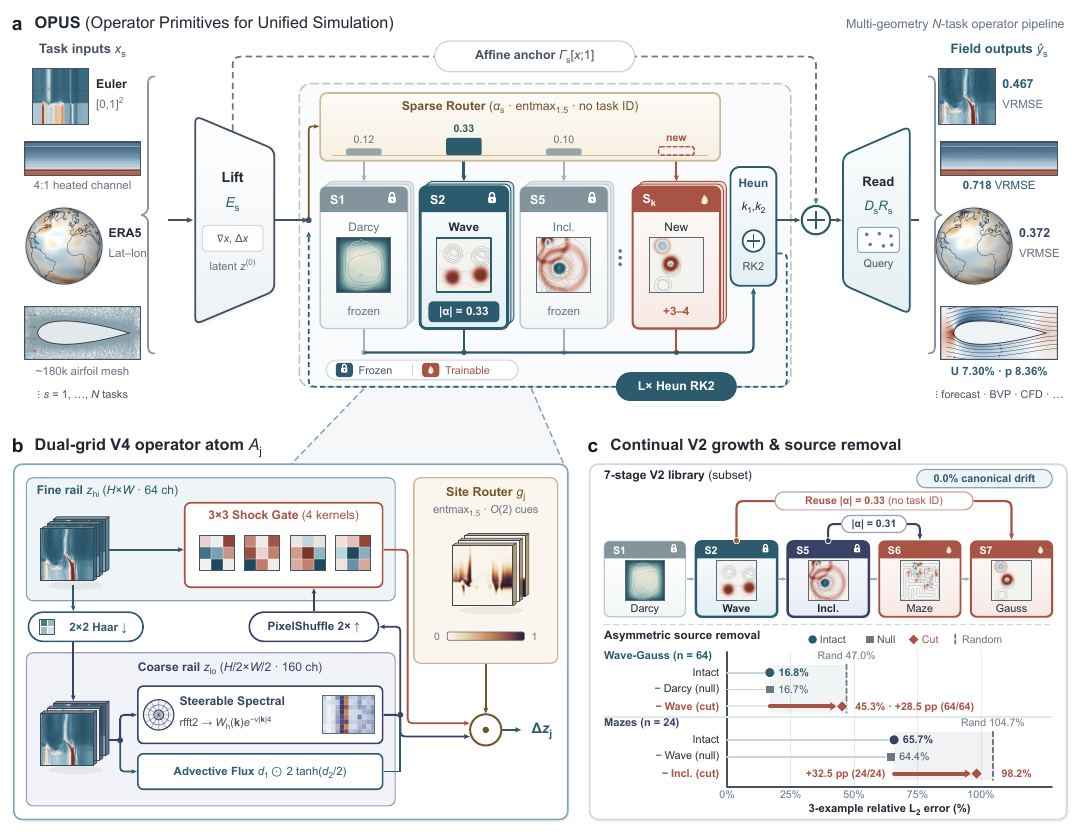}
  \caption{\textbf{OPUS (Operator Primitives for Unified Simulation) for continual multi-geometry physics.} \textbf{a}, Open-ended \(N\)-task multi-geometry overview mapping arbitrary task inputs \(x_s\) across diverse domains (illustrated with V4 Euler shocks on \([0,1]^2\), \(4:1\) heated Rayleigh--B\'enard channels, global ERA5 atmosphere on a latitude--longitude grid shown in globe projection, and V3 \(\sim\)180k-node AirfRANS airfoil meshes) through a geometry lift (\(E_s\), stage stem in V2), an unrolled \(L\)-step indexed operator library of frozen prior stages (padlock) and newly appended trainable blocks (flame) combined by a router (V2 block weights \(\alpha_m\) fitted from paired support without a task ID before enrollment, and V4 spatial \texttt{entmax-1.5} weights over stage-eligible blocks), and a readout (ridge in V2; anchored residual in V3/V4) onto diverse field outputs (rollouts, steady-state BVPs, scattering and CFD). \textbf{b}, Magnified view of one \textbf{V4} dual-resolution operator atom \(A_j\) coupling a \(128\times128\) Euler shock fine rail (\(z_{\mathrm{hi}}\)) and its \(2\times2\) Haar-packed \(64\times64\) coarse rail (\(z_{\mathrm{lo}}\)) via \(2\times\) PixelShuffle, with four learned \(3\times3\) fine shock kernels, an \(8\times12\) steerable spectral gain, bounded advective local flux and per-site Euler \texttt{entmax-1.5} routing maps. \textbf{c}, Representative stages from the seven-stage \textbf{V2} library (exact \(0.0\%\) canonical replay drift; per-source mean absolute block weights \(\overline{|\alpha|}_{\mathrm{src}}=0.325\) from S2 Wave-Layer on S7 Wave-Gauss and \(\overline{|\alpha|}_{\mathrm{src}}=0.305\) from S5 Acoustic Inclusions on S6 Acoustic Mazes, fitted without target physics labels) and paired asymmetric source-removal interventions on Wave-Gauss (\(n=64\): \(+28.47\)~pp when Wave-Layer is removed versus \(-0.06\)~pp for Darcy) and Acoustic Mazes (\(n=24\): \(+32.50\)~pp when the Acoustic Inclusions source is removed versus \(-1.36\)~pp for Wave-Layer; full paired intervals in Fig.~\ref{fig:interventions}).}
  \label{fig:library}
\end{figure}

Figure~\ref{fig:library} summarizes the OPUS library structure across its V2 and V4 instantiations. Each training stage freezes all previously acquired input stems and operator blocks, appends new trainable blocks tagged with their source stage, and stores the task's normalization statistics, eligible block indices \(J_s\), mixing weights and readout. Because earlier parameters are never overwritten and stored routes restrict evaluation to the stage's recorded indices, canonical replay reproduces stored predictions identically. In V3 and V4, a cap on \(|J_s|\) (up to six or eight earlier blocks plus four new ones) bounds the blocks evaluated per unrolled update while storage grows linearly with enrollment; V2 adaptation mixes all frozen blocks.

\paragraph{V2 model and adaptation methods (primary intervention testbed).}
In V2, every input field channel is standardized on its support set using both linear (mean/standard deviation) and robust signed-log statistics, then expanded into 13 parameter-free spatial features (standardized value, signed-log value, two centered first differences, discrete Laplacians at offsets 1, 2 and 4 grid cells, and six Cartesian/sinusoidal coordinates). Each enrolled stage owns a lightweight channel-equivariant convolutional stem (\texttt{StageStem}, latent width \(d=80\), context width \(40\), \(K_r=4\) channel roles, RMS-normalized) and appends 2--4 operator blocks drawn from three families: low-rank 2D FFT spectral multipliers (\texttt{SpectralAtom}, rank 20, 12 modes), face-gated directional flux and dilated convolutions (\texttt{FluxSpecAtom}, dilation 2 or 4 per block), and pointwise gated trilinear interactions (\texttt{GatedReactionAtom}). Latent states evolve over \(L=8\) forward-Euler residual steps of size \(1/L\), \(z^{(\ell+1)} = z^{(\ell)} + L^{-1}\sum_{m}\alpha_{\ell,m} A_m(z^{(\ell)}, h_{\mathrm{ctx}})\), followed by a linear ridge readout over \([z^{(L)}, x_{\mathrm{lin}}, x_{\mathrm{slog}}, 1]\).

During pre-enrollment few-example adaptation (three labeled support pairs, no target physics label), all frozen stems and blocks remain fixed; we fit a step-shared mixing coefficient \(\alpha_m\) per frozen block alongside the low-dimensional channel-role matrices and stem mixing weights for 55 Adam steps (learning rate 0.06, \(\ell_1\) penalty \(2\times10^{-3}\) on block and stem mixing weights, \(\ell_2\) penalty \(10^{-3}\) on these mixing weights and the channel-role matrices) by minimizing leave-one-trajectory-out (LOTO) ridge prediction error on the three support pairs, then select the ridge regularization parameter over \(\{10^{-4},\ldots,1.0\}\) by LOTO error and refit the final linear readout on all three pairs. For each source-removal intervention, we mask out both the source stage's stem and all operator blocks acquired at that stage before repeating the identical optimization on the exact same support draws and held-out query fields. Full architecture specifications, per-stage training schedules and bootstrap procedures are detailed in Supplementary Section~\ref{sec:s1}.

\paragraph{V3 and V4 routing variants.}
The routing mechanism evolves across versions. V2 fits step-dependent mixing weights during permanent enrollment and step-shared mixing weights during three-example LOTO adaptation. V3 ranks frozen blocks by how well each block explains labeled target residuals, then learns step- and channel-wise gates alongside an affine anchor plus learned residual, \(\widehat{y}_s(x) = \Gamma_s[x;1] + D_s R_s(\Phi_s(x;J_s))\). V4 replaces single-resolution blocks with dual-resolution coupled blocks (64 fine channels Haar-packed to \(4\times64\) coarse channels and linearly projected to 160, returned by \(2\times\) PixelShuffle), evolved with two Heun (RK2) macro-steps weighted by per-site \texttt{entmax-1.5} spatial routing \cite{entmax}. Supplementary Fig.~\ref{fig:operator-detail} traces V4's block computation and measured stage weights, and Supplementary Fig.~\ref{fig:task-atlas} pairs V2 and V4 forecasts with their reference fields.

\section{Source-specific causal interventions in few-example transfer}
\label{sec:interventions}

\begin{figure}[!t]
  \centering
  \includegraphics[width=\linewidth,height=4.70in,keepaspectratio]{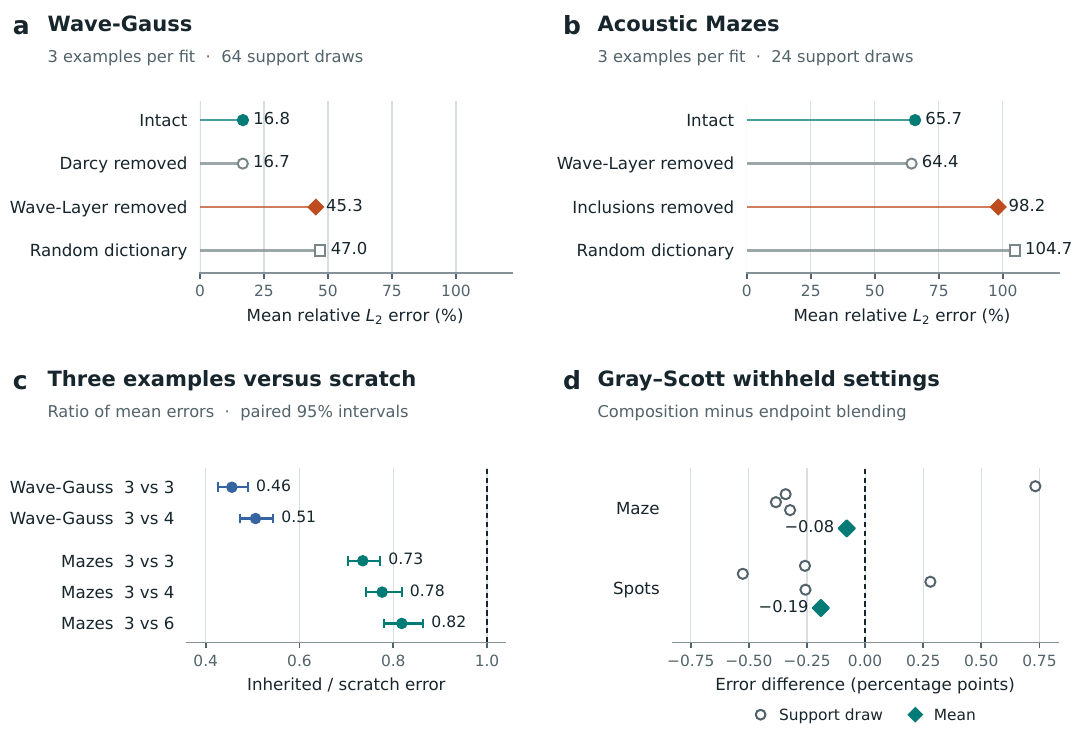}
  \caption{\textbf{Selective reuse before enrollment and withheld-parameter composition in V2.} \textbf{a,b}, Three-example adaptations on Wave-Gauss (64 draws from the pre-Gauss Stage-6 checkpoint) and Acoustic Mazes (24 draws from the pre-Maze Stage-5 checkpoint) with the indicated source's input stem and operator blocks removed together. \textbf{c}, Ratios of mean errors for inherited and scratch fits; bars show 95\% paired bootstrap confidence intervals (5,000 resamples) over support draws on fixed query fields. \textbf{d}, Each point compares in-loop operator composition with an endpoint-output affine blend on one withheld Gray--Scott support draw.}
  \label{fig:interventions}
\end{figure}

Figure~\ref{fig:interventions} presents the pre-enrollment intervention tests on V2, evaluated using frozen checkpoints saved immediately prior to each target's enrollment (Stage 6 for Wave-Gauss; Stage 5 for Acoustic Mazes). Support examples and query fields come from disjoint trajectories; Wave-Gauss interventions are scored on a fixed 144-pair subset of the 240 held-out query pairs in Table~\ref{tab:v2full}, and Acoustic Mazes on all 48. Reported percentage-point differences are computed from unrounded error means.

On \textbf{Wave-Gauss}, three labeled examples adapted on the inherited library achieved 16.63\% mean relative \(L_2\) error across 32 paired support draws versus 36.49\% for scratch training on identical supports, winning on all 32 draws (ratio of mean errors 0.456, 95\% paired bootstrap CI 0.427--0.490; Fig.~\ref{fig:interventions}c). Across the full 64-draw source-removal audit on the same pre-Gauss checkpoint, intact adaptation reached 16.80\% mean error, assigning its largest per-source mean absolute block weight, \(\overline{|\alpha|}_{\mathrm{src}} = |M_{\mathrm{src}}|^{-1}\sum_{m\in M_{\mathrm{src}}}\overline{|\alpha_m|} = 0.325\) (\(M_{\mathrm{src}}\): blocks acquired at the source stage; \(\overline{|\alpha_m|}\): mean of \(|\alpha_m|\) over support draws), to the earlier Wave-Layer source (versus 0.118 on Darcy and 0.096 on Acoustic Inclusions). Disabling the Wave-Layer stem and blocks raised error by 28.47~pp to 45.27\% (intact-to-removed ratio 0.371, 95\% paired bootstrap CI 0.347--0.399, 64 of 64 draws worse), nearly matching a randomly initialized library of identical layout (47.04\%, intact-to-random ratio 0.357, 95\% paired bootstrap CI 0.332--0.387, intact lower on 64 of 64 draws). Disabling the Darcy source changed error by \(-0.06\)~pp to 16.74\% (Fig.~\ref{fig:interventions}a).

On \textbf{Acoustic Mazes} (24 paired draws on the pre-Maze Stage-5 checkpoint), intact three-example adaptation reached 65.73\% mean error and assigned its largest per-source mean absolute block weight (\(\overline{|\alpha|}_{\mathrm{src}} = 0.305\)) to the earlier Acoustic Inclusions source (versus 0.129 on Wave-Layer). Disabling Acoustic Inclusions raised error by 32.50~pp to 98.24\% (intact-to-removed ratio 0.669, 95\% paired bootstrap CI 0.638--0.707, 24 of 24 draws worse; random library 104.67\%), whereas disabling Wave-Layer, the critical donor for Wave-Gauss, yielded 64.38\% (\(-1.36\)~pp; Fig.~\ref{fig:interventions}b). Three inherited examples also outperformed six scratch examples on 22 of 24 nested draws (80.33\% scratch mean, ratio 0.818, 95\% paired bootstrap CI 0.780--0.864). Thus, even where absolute few-example accuracy remains poor, the intervention isolates a large source-specific contribution.

Three qualifications bound these results. First, the two experiments form paired asymmetric source-removal contrasts rather than a complete \(2\times2\) double dissociation (Acoustic Inclusions removal was not scored on Wave-Gauss, where Darcy served as the comparison removal). Second, each removal disables the source's input stem and operator blocks together, so the experiment locates reusable information at the source-stage bundle level rather than separating stem compatibility from spatial operator weights. Third, while the scratch comparator receives the exact same 13-feature differential jet, it uses a separately optimized end-to-end training routine with different parameter counts and batching, so comparisons measure label efficiency rather than equal-compute or equal-capacity speedups. The Wave-Gauss six-label scratch run used differently indexed support sets and is excluded from the paired ratio panel.

\section{Full-data accuracy, retention and regime identification}
\label{sec:full}

\begin{table}[!t]
  \centering
  \caption{\textbf{Permanent V2 enrollment: field pair counts and per-task mean relative \(L_2\) field error.} All available pairs in each staged training split were used for permanent V2 enrollment. The final two columns compare continual growth against an independently trained single-task scratch model on the identical split.}
  \label{tab:v2full}
  \small
  \begin{tabular}{@{}lrrrr@{}}
    \toprule
    Stage & Training pairs & Query pairs & Continual (\%) & Scratch (\%)\\
    \midrule
    1\quad Darcy & 9,000 & 1,000 & 2.60 & 2.60\\
    2\quad Wave-Layer & 10,212 & 240 & 0.79 & 0.74\\
    3\quad Directional diffusion & 50 & 30 & 8.71 & 9.30\\
    4\quad Buoyant flow & 64 & 48 & 20.82 & 20.88\\
    5\quad Acoustic Inclusions & 2,798 & 48 & 7.02 & 6.79\\
    6\quad Acoustic Mazes & 1,600 & 48 & 20.98 & 19.59\\
    7\quad Wave-Gauss & 10,212 & 240 & 0.95 & 0.92\\
    \bottomrule
  \end{tabular}
\end{table}

Table~\ref{tab:v2full} reports full-data permanent enrollment across the seven V2 tasks registered onto a shared \(32\times32\) grid: stationary Darcy flow from PDEBench \cite{pdebench}, Wave-Layer and Wave-Gauss from Poseidon \cite{poseidon}, anisotropic directional diffusion from APEBench \cite{apebench}, incompressible buoyant flow from PDEArena \cite{pdearena}, and Acoustic Inclusions and Acoustic Mazes from \emph{The Well} \cite{well}. Relative field error is
\begin{equation}
  \varepsilon = \frac{1}{NC}\sum_{i=1}^{N}\sum_{c=1}^{C}\frac{\lVert\widehat{y}_{i,c}-y_{i,c}\rVert_2}{\lVert y_{i,c}\rVert_2},
  \label{eq:rell2}
\end{equation}
with Euclidean norms taken over grid points. Wave-Layer and Wave-Gauss share identical 10,212/240 train/query splits by design (each drawn from Poseidon's 10,512-trajectory archive, with 60 validation trajectories held separately), and both predict frame 2 from frame 0 (a shorter lead than Poseidon's frame-14 evaluation \cite{poseidon}). Each scratch model has one stem and four blocks, whereas the inherited library accumulates up to seven stems and 21 blocks (2.65M parameters across seven stages). Three of seven tasks fall below 5\% mean error, and scratch achieves slightly lower error on four of seven tasks. Continual pretraining therefore yields no systematic full-data advantage on V2; its benefit concentrates in scarce-example adaptation.

\paragraph{What survives a change of support?}
Canonical replay using stored task settings yielded zero drift on every enrolled stage, as guaranteed by construction. Re-identifying an earlier task from four freshly sampled support pairs (without stored routes) behaved differently: between the stage of acquisition and Stage 7, fresh-support Wave-Layer error rose from 3.45\% to 5.16\%, whereas Darcy error fell from 17.97\% to 10.97\%. Canonical replay and fresh-support re-identification test distinct deployment regimes and should not be conflated.

\paragraph{Identifying a known response.}
On 42 standardized eight-example support banks across the seven enrolled tasks, leave-one-source-out residual attribution identified the known owner on 42 of 42 banks (100\%). However, an input-only moment classifier also achieved 95.2\%, showing that input field statistics alone provide a strong identification cue in multi-task archives. Two additional probe tests supplied identical inputs to distinct sources and generated targets using the model's own stored readouts, providing a sanity check of stored responses rather than independent evidence of physical identification. In an independent synthetic control applying fixed Fourier filters to shared random inputs (3-way chance 33.3\%, 60 banks), the largest learned source weight selected the matching physics owner on 47 of 60 banks (78.3\%), while removal-based identification reached 66.7\%.

\paragraph{Predicting at withheld parameter settings.}
Four Gray--Scott parameter endpoints (gliders, spirals, worms, bubbles from \emph{The Well} \cite{well}) were enrolled as additional stages atop the Stage-7 V2 library (outside Table~\ref{tab:v2full}) and frozen, and three labeled examples were supplied from two withheld regimes: interior \emph{maze} and exterior \emph{spots} (the latter outside the convex hull of the four endpoints in the feed--kill plane; Fig.~\ref{fig:interventions}d). Across four draws per setting, in-loop operator composition achieved 6.37\% error on \emph{maze} and 5.51\% on \emph{spots}, compared with 6.45\% and 5.70\% for an affine blend of endpoint predictions (winning three of four draws per setting). With so few draws and margins below 0.2~pp, an advantage of in-loop composition over endpoint blending is not established.

\section{Transfer to unseen simulation families and ocean anomalies}
\label{sec:unseen}

\begin{figure}[!t]
  \centering
  \includegraphics[width=\linewidth,height=3.95in,keepaspectratio]{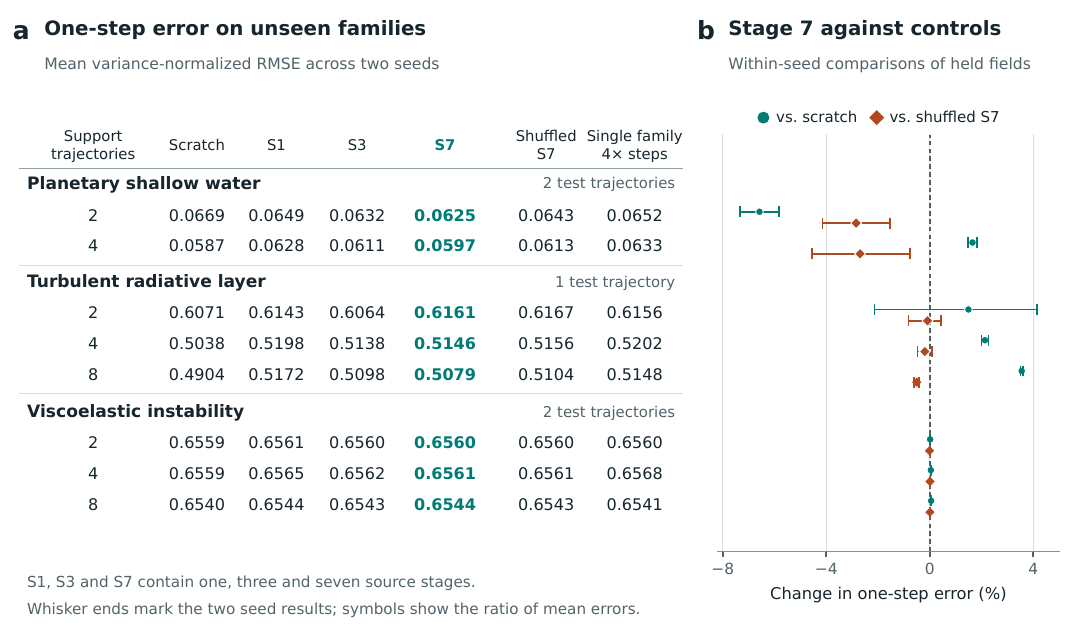}
  \caption{\textbf{One-step transfer to unseen families from \emph{The Well} in V3.} Each support example is a full trajectory supplying multiple labeled field transitions. Frozen checkpoints are evaluated alongside scratch and scrambled controls at the same target budget. \textbf{a}, Six methods (scratch, Stage-1, Stage-3, Stage-7, scrambled Stage-7 and a single-family \(4\times\)-step control) across eight family-and-budget combinations (planetary shallow water at 2 and 4 trajectories; turbulent radiative layer and viscoelastic instability at 2, 4 and 8 trajectories). \textbf{b}, Learned seventh-stage error relative to scratch and scrambled controls; whisker ends show the two individual seed outcomes. Evaluation uses one radiative test trajectory and two per other family. Library content, capacity and training history vary together.}
  \label{fig:unseen}
\end{figure}

V3 sequentially enrolled five sampled simulation families from \emph{The Well} \cite{well}, followed by ERA5 atmosphere \cite{era5,wb} and NOAA OISST sea surface temperature \cite{oisst} (Supplementary Table~\ref{tab:s1}; 28 blocks, 14.77M parameters). On shear flow, tenth-step variance-normalized root-mean-square error (VRMSE) reached 0.2564 for continual V3 versus 0.4480 for scratch (persistence 0.7795). On Rayleigh--B\'enard convection, as on active matter, both models were worse than persistence at step ten (2.0106 continual, 1.8821 scratch, 0.8291 persistence), underscoring that multi-step rollout accuracy depends strongly on physical regime and horizon.

Figure~\ref{fig:unseen} summarizes frozen transfer to three unseen families from \emph{The Well} by fitting a lightweight projection and readout on new trajectories. With two planetary shallow-water trajectories, one-step VRMSE decreased monotonically from 0.0649 (Stage-1 library) to 0.0632 (Stage 3) and 0.0625 (Stage 7), improving on scratch (0.0669) and a scrambled Stage-7 library (0.0643). At four trajectories, however, scratch improved to 0.0587, below the learned library's 0.0597. On the turbulent radiative layer with eight trajectories (one test trajectory), Stage 7 (0.5079) edged out scrambled (0.5104) while scratch reached 0.4904; on viscoelastic instability (two trajectories), learned and scrambled both scored 0.6560, versus 0.6559 for scratch.

\paragraph{What improved the ocean forecast?}
On the five-day NOAA OISST anomaly task \cite{oisst}, permanent V3 enrollment gave 61.12\% relative error in 2023, worse than its matched scratch counterpart (59.07\%), its Stage-7 affine anchor (49.76\%) and persistence (50.15\%). In a separate chronological fit, we trained an early-stopping adapter on the frozen six-stage library using 2018--2021 data, selected checkpoints on 2022 and computed the anomaly climatology over 2018--2022; 2023 error then improved to 49.41\%, but scrambled six-stage blocks gave 49.18\%, scratch gave 49.02\%, the chronological affine anchor gave 49.77\% and persistence gave 50.15\%. Under the chronological protocol, the scrambled and scratch controls both had lower 2023 error than the learned adapter, showing no detectable pretraining advantage over controls; moreover, the 2018--2022 anomaly climatology included the 2022 validation year and the 2023 cohort had been scored during earlier development.

\section{Sequential forecasts on square, channel and latitude--longitude domains}
\label{sec:v4}

\begin{figure}[!t]
  \centering
  \includegraphics[width=\linewidth,height=5.65in,keepaspectratio]{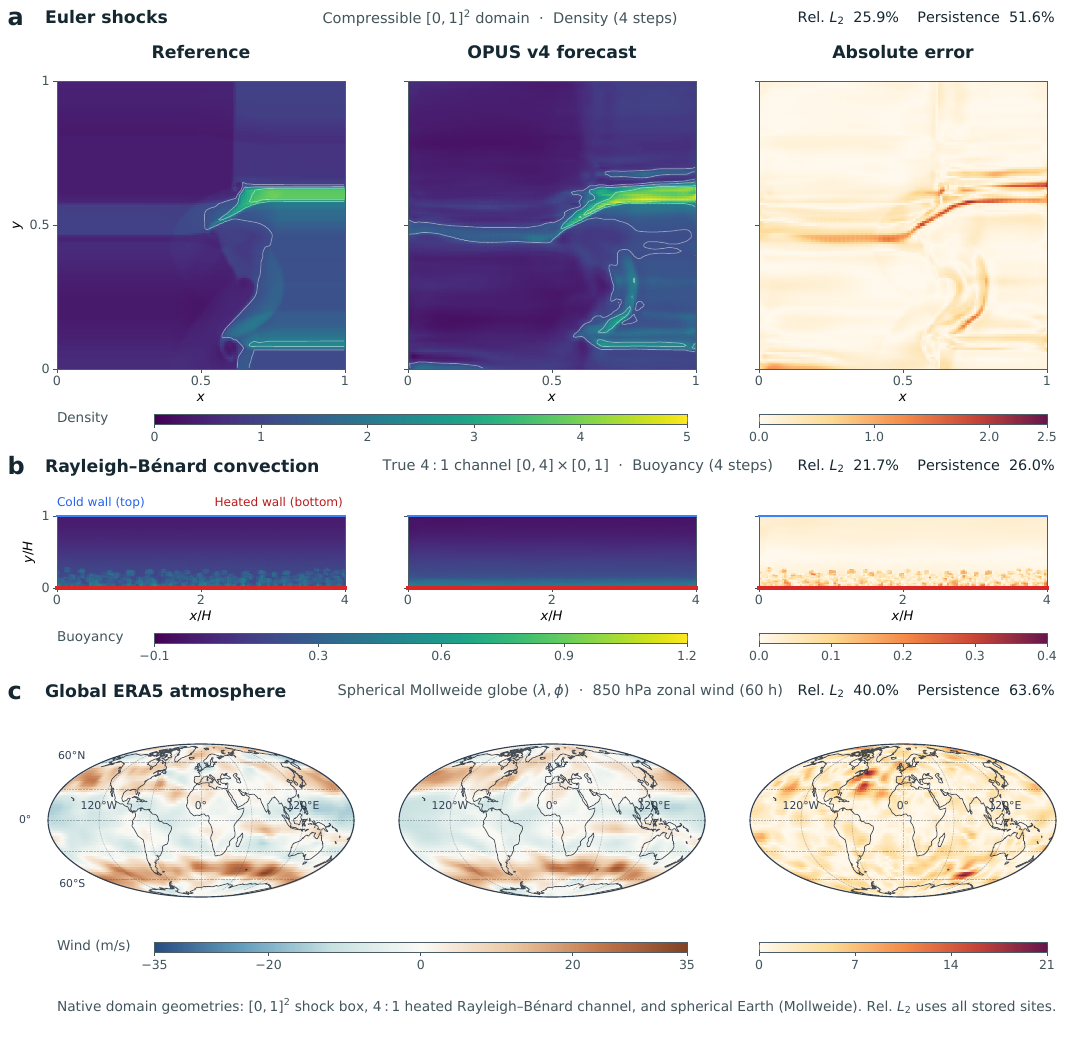}
  \caption{\textbf{Held-out forecasts from the three-stage V4 checkpoint rendered on their physical domains.} Reference, forecast and pointwise absolute error are shown for \textbf{a}, compressible Euler shock density on \([0,1]^2\) with iso-density contours (4 steps, resized to \(128\times128\)); \textbf{b}, Rayleigh--B\'enard buoyancy in its \(4:1\) heated channel \([0,4]\times[0,1]\) between cold top and heated bottom walls (4 steps, resized to \(128\times128\)); and \textbf{c}, global ERA5 850~hPa zonal wind shown in a Mollweide projection with coastlines (10 six-hour steps, 60~h; \(32\times64\) WeatherBench~2 source upsampled to \(64\times128\)). Example errors describe one held-out trajectory per domain.}
  \label{fig:fields}
\end{figure}

To test whether a dual-resolution operator library scales beyond \(32\times32\) squares, we evaluated V4 across compressible Euler shocks (natively \(512\times512\), resized to \(128\times128\)), Rayleigh--B\'enard convection (\(4:1\) channel, natively \(512\times128\), represented on a \(128\times128\) array over \([0,4]\times[0,1]\)) and global ERA5 atmosphere (\(32\times64\) WeatherBench~2 \cite{wb,era5} equiangular latitude--longitude grid bilinearly upsampled to \(64\times128\), with VRMSE computed on the \(64\times128\) grid; Fig.~\ref{fig:fields}). In single-seed, single-target 1,600-update comparisons against FNO2D \cite{fno}, U-Net \cite{unet} and V3 (Supplementary Table~\ref{tab:s2}), rollout-trained V4 achieved the lowest tenth-step VRMSE on both Euler shocks (0.4591 versus 0.4626 U-Net, a 0.8\% margin, and 0.7091 FNO2D) and 60-hour global ERA5 (0.3474 versus 0.3874 U-Net and 0.4426 FNO2D), while one-step-trained V4 led on Rayleigh--B\'enard (0.8462 versus 0.8546 U-Net).

\begin{table}[!t]
  \centering
  \caption{\textbf{Tenth-step VRMSE after sequential learning (shocks, convection, atmosphere).} Each model's mean weights the three datasets equally. Sequential U-Net overwrites shared weights without rehearsal, whereas V4 appends four blocks per stage and uses stored routes.}
  \label{tab:v4seq}
  \small
  \begin{tabular}{@{}lrrrr@{}}
    \toprule
    Final model & Shocks & Convection & Atmosphere & Mean\\
    \midrule
    U-Net, 2.25M parameters (overwritten) & 5.8317 & 2.4416 & 0.3978 & 2.8904\\
    V4, 4.42M parameters (stored routes) & 0.4671 & 0.7176 & 0.3721 & 0.5189\\
    \bottomrule
  \end{tabular}
\end{table}

Table~\ref{tab:v4seq} contrasts catastrophic forgetting under naive sequential fine-tuning of a shared-weight U-Net with V4 continual growth (mean tenth-step VRMSE 2.8904 versus 0.5189, with exact canonical replay on four saved inputs per earlier V4 stage). The U-Net's tenth-step VRMSE degraded from 0.4750 at acquisition to 5.8317 on Euler shocks and from 0.8142 to 2.4416 on convection. Because V4 expands to 4.42M parameters and retains stage routes, Table~\ref{tab:v4seq} compares two deployment policies rather than equal-memory continual baselines; three independently archived four-block V4 specialists (\(\approx 4.5\)M total parameters) would likewise guarantee exact replay and are needed to isolate cross-stage reuse from parameter isolation. Sequential V4 convection error (0.7176) is lower than every single-target model in Supplementary Table~\ref{tab:s2} (best 0.8462); however, the sequential run evaluates eight active blocks (four frozen Euler plus four new convection blocks) under a two-phase alignment schedule rather than four single-phase blocks, so we do not attribute that gap to cross-stage reuse without a matched capacity and schedule control. On two held-out families after three-stage V4 acquisition, frozen V4 improved over scrambled V4 on two planetary shallow-water trajectories (0.0562 versus 0.0585 at step 1; 0.3895 versus 0.4011 at step 10), but learned blocks trailed scrambled blocks at step 10 with four shallow-water trajectories (0.3542 versus 0.3314) and on the turbulent radiative layer (0.9382 versus 0.7920 with two trajectories; 0.9135 versus 0.8385 with four), and end-to-end V4 scratch won all four family-and-budget rows (Supplementary Table~\ref{tab:s3}).

\section{Unstructured airfoil meshes: limits of frozen reuse}
\label{sec:airfoil}

\begin{figure}[!t]
  \centering
  \includegraphics[width=\linewidth,height=4.75in,keepaspectratio]{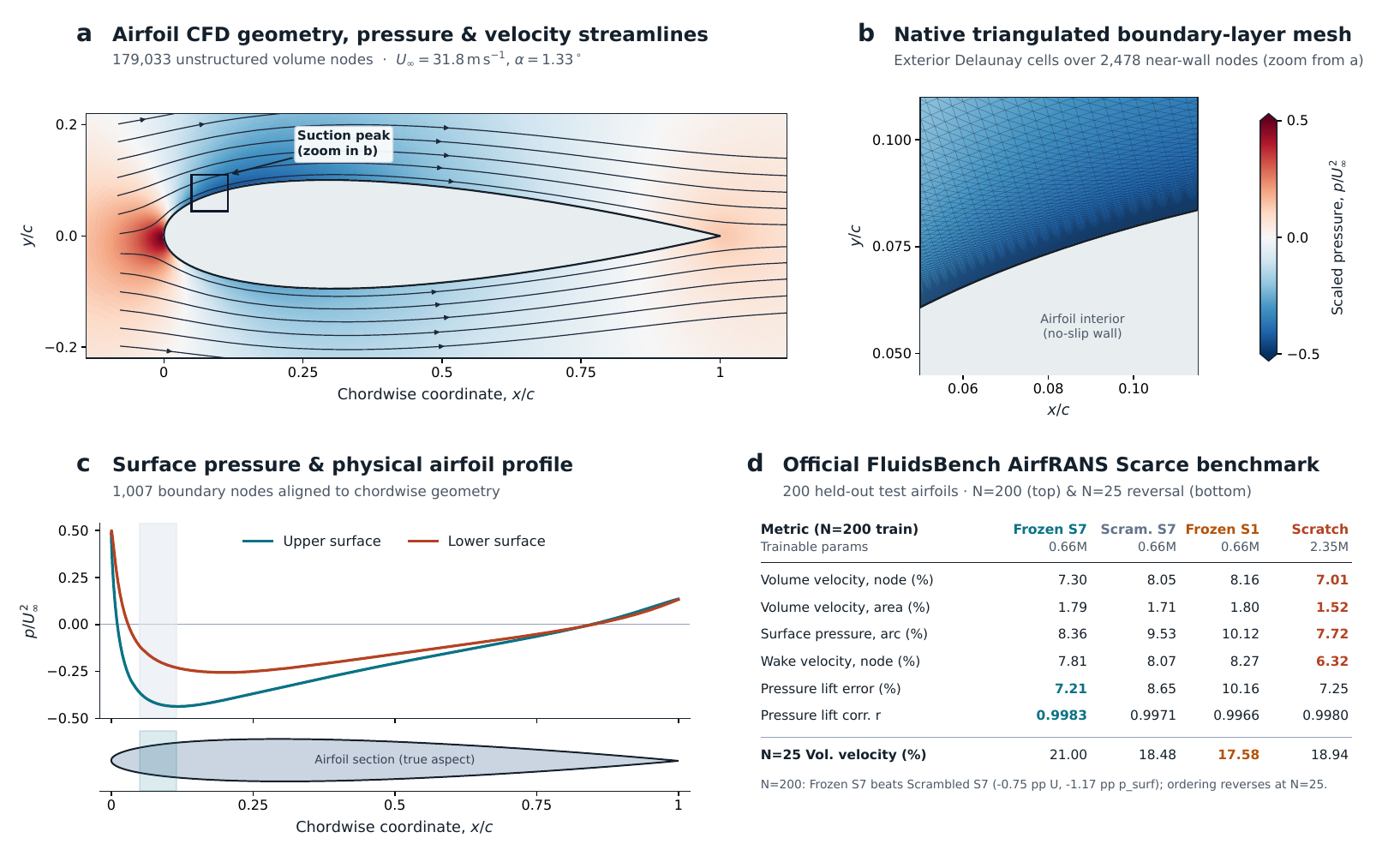}
  \caption{\textbf{Unstructured AirfRANS CFD geometry, near-wall Delaunay visualization, and scarce-split benchmark.} \textbf{a}, Scaled CFD pressure \(p/U_\infty^2\) and streamlines from the native velocity vector field across 179,033 unstructured volume nodes around one displayed airfoil. \textbf{b}, Magnified exterior Delaunay triangulation reconstructed over 2,478 near-wall nodes along the curved upper airfoil wall. \textbf{c}, Upper- and lower-surface pressure profiles across 1,007 boundary nodes aligned above the airfoil cross-section. \textbf{d}, Four-arm evaluation at the official AirfRANS Scarce split (\(N=200\)) and a custom 25-case training subset (\(N=25\)), both scored on the same 200 official test airfoils.}
  \label{fig:airfoil}
\end{figure}

AirfRANS supplies steady simulated velocity and pressure fields on curved, body-fitted unstructured meshes \cite{air,fluids}. A trainable geometry-informed interface maps unstructured airfoil nodes onto a stretched \(128\times128\) latent grid, applies six manually selected frozen V3 operator blocks (one from each of the first six stages of the seven-stage V3 checkpoint) and decodes predictions back onto the native volume and boundary nodes (Fig.~\ref{fig:airfoil}). Because block indices were fixed by the experimenter, this benchmark isolates representation value inside a mesh adapter rather than automatic routing.

On the official AirfRANS Scarce split (\(N=200\) training, 200 test airfoils), node-weighted domain velocity relative \(L_2\) error (evaluated across all native volume nodes, distinct from the original AirfRANS MSE and Spearman leaderboard table \cite{air}) was 7.30\% for the frozen learned library (0.662M trainable parameters), 8.05\% for the same interface with scrambled frozen blocks, 8.16\% for the Stage-1-only library and 7.01\% for a GINO scratch comparator \cite{gino} (2.35M trainable parameters). Arc-length-weighted surface pressure relative \(L_2\) errors followed the same ordering: 8.36\%, 9.53\%, 10.12\% and 7.72\%, respectively. Recorded wall-clock training times on a single NVIDIA GH200 GPU were 304~s for the frozen-library adapter and 251~s for scratch, so fewer trainable parameters did not reduce elapsed training time. On a custom 25-airfoil subset scored on the same 200 test airfoils, the velocity ordering changed and the frozen learned library ranked last: frozen learned 21.00\%, scrambled 18.48\%, Stage-1 17.58\% and scratch 18.94\%. All four arms represent single training runs; multi-seed variance remains unmeasured.

\section{Discussion, limitations and reproducibility}
\label{sec:discussion}

Source-specific intervention provides this study's clearest mechanistic evidence. On V2 Wave-Gauss and Acoustic Mazes, paired source removal shows that, in these two near-transfer cases, three-example adaptation depends selectively on Wave-Layer for Wave-Gauss and on Acoustic Inclusions for Acoustic Mazes; removing the respective comparison sources (Darcy and Wave-Layer) changed mean error by \(-0.06\) and \(-1.36\)~pp. This yields an operational, testable definition of selective reuse in continual operator learning without attributing symbolic equations to hidden features. At the same time, the full-data V2 comparison and the V3, V4 and AirfRANS audits show sharp empirical boundaries: full-data continual fits show no systematic advantage over scratch, frozen transfer to unseen families mostly trails end-to-end scratch (the only one-step exception, two-trajectory V3 shallow water, reverses at four trajectories), and learned frozen representations beat scrambled controls on 200 airfoils yet rank last under 25-case scarcity.

Several limitations define clear next steps. First, our V2 intervals resample support draws on fixed query fields from a single trained seven-stage library rather than multiple random library seeds; moreover, each source removal disables stems and operator blocks jointly, and the two target tasks use different comparison removals rather than a full source-by-target ablation matrix. Second, V2, V3 and V4 use different architectures, metrics and training histories rather than a single unified flagship model. Third, the parameter-free 13-feature jet encodes a useful spatial prior, stored-route replay is exact by construction rather than through weight-space consolidation, and the sequential U-Net comparison lacks equal-memory continual baselines (such as EWC, replay or archived specialists). Fourth, both Wave-Layer \(\to\) Wave-Gauss and Acoustic Inclusions \(\to\) Acoustic Mazes represent near-transfer within related PDE families; demonstrating selective reuse across genuinely distant equation families remains open.

Two controlled follow-up experiments would directly test whether these findings generalize: (1) retraining the expanded library across multiple seeds and running the complete source-by-target removal matrix (separating stem-only from block-only removals) alongside equal-budget random-library and specialist baselines; and (2) evaluating repeated multi-seed AirfRANS and unseen-family runs on a single scaled continual checkpoint.

\paragraph{Acknowledgements.}
This work used the Delta and DeltaAI systems at the National Center for Supercomputing Applications [awards OAC 2005572 and OAC 2320345] through allocation CIS240093 from the Advanced Cyberinfrastructure Coordination Ecosystem: Services \& Support (ACCESS) program, which is supported by National Science Foundation grants \#2138259, \#2138286, \#2138307, \#2137603, and \#2138296.

\paragraph{Data and code availability.}
V2 datasets derive from PDEBench (Darcy) \cite{pdebench}, Poseidon (Wave-Layer and Wave-Gauss) \cite{poseidon}, APEBench (directional diffusion) \cite{apebench}, PDEArena (buoyant flow) \cite{pdearena} and \emph{The Well} (Acoustic Inclusions, Acoustic Mazes and Gray--Scott) \cite{well}; V3, V4 and AirfRANS datasets derive from \emph{The Well} \cite{well}, WeatherBench~2 / ERA5 \cite{wb,era5}, NOAA OISST \cite{oisst} and AirfRANS / FluidsBench \cite{air,fluids}. All processed arrays, splits, plot-ready figure packets and normalization statistics needed to reproduce the reported numbers are available from the corresponding author on reasonable request. The OPUS implementation, training and evaluation scripts, baseline adapters and pretrained checkpoints will be made available upon publication.

\paragraph{Use of large language models.}
Large language models were used to assist with editing the text of the manuscript and with debugging analysis code. All experimental designs, results, interpretations and conclusions are the authors' own, and all text was reviewed and revised by the authors. No large language model is an author of this work.

\paragraph{Author contributions.}
S.R., Y.P.L., K.K., J.P.Y., S.C. and S.B.A. contributed to the conception, execution and writing of this work.

\paragraph{Competing interests.}
The authors declare no competing interests.

\clearpage
\appendix
\setcounter{figure}{0}
\setcounter{table}{0}
\setcounter{equation}{0}
\renewcommand{\thefigure}{S\arabic{figure}}
\renewcommand{\thetable}{S\arabic{table}}
\renewcommand{\theequation}{S\arabic{equation}}
\renewcommand{\thesection}{S\arabic{section}}
\begin{center}
  {\Large\bfseries Supplementary Material:\\[2pt] Methods, Extended Benchmarks and Visual Atlas}
\end{center}
\vspace{4pt}

\section{Detailed V2 architecture, training and source-removal protocol}
\label{sec:s1}

\paragraph{Input representation and stage stems.}
For each available field channel on the registered \(32\times32\) grid, V2 computes support-only linear (mean and standard deviation) and robust signed-log (median and interquartile range) normalization statistics so that query fields never influence scaling. From the paired linear and signed-log channels, a parameter-free differential jet forms 13 features per input channel: the linearly standardized field, the signed-log standardized field, two centered first differences, discrete Laplacians at offsets of 1, 2 and 4 grid cells (with edge-replicated padding), and six Cartesian and sinusoidal coordinate maps \((x,y,\sin\pi x,\cos\pi x,\sin\pi y,\cos\pi y)\). Each enrolled stage \(s\) owns a compact convolutional stem (\texttt{StageStem}, two \(3\times3\) SiLU convolutions lifting \(13\to40\to80\) channels followed by \(1\times1\) projections into \(K_r=4\) channel roles for latent state \(z\in\mathbb{R}^{80}\) and spatial context \(h_{\mathrm{ctx}}\in\mathbb{R}^{40}\), each RMS-normalized). Task-specific role matrices \(a_{\mathrm{in}}, b_{\mathrm{in}}\in\mathbb{R}^{4\times C_{\mathrm{in}}}\) and stem mixing weights \(w_{\mathrm{stem}}\) combine the active stage stems into the initial latent state \(z^{(0)}\) and context \(h_{\mathrm{ctx}}\).

\paragraph{Operator block families and unrolled dynamics.}
Each stage freezes all prior stems and blocks and appends 2--4 new RMS-normalized operator blocks from three families: (i) \texttt{SpectralAtom}, a low-rank 2D FFT spectral multiplier (rank 20, 12 retained Fourier modes in each positive and negative frequency quadrant) modulated by a context gate \(1 + 0.5\tanh(\mathrm{Conv}_{1\times1}(h_{\mathrm{ctx}}))\); (ii) \texttt{FluxSpecAtom}, four sigmoid-gated directional neighbor differences concatenated with local \(3\times3\) and dilated \(3\times3\) (dilation 2 or 4 per block) depthwise convolutions over 40 hidden channels; and (iii) \texttt{GatedReactionAtom}, a pointwise gated trilinear interaction over 48 hidden channels. The seven enrolled stages append, in order: Darcy (4 blocks: 2 spectral, 1 flux, 1 reaction; 28 epochs, batch 64), Wave-Layer (3 blocks: spectral, flux, reaction; 28 epochs, batch 64), Directional diffusion (2 blocks: spectral, flux; 80 epochs, batch 16), Buoyant flow (2 blocks: flux, reaction; 80 epochs, batch 16), Acoustic Inclusions (4 blocks: 2 flux, 1 spectral, 1 reaction; 36 epochs, batch 48), Acoustic Mazes (3 blocks: 2 flux, 1 spectral; 42 epochs, batch 48) and Wave-Gauss (3 blocks: spectral, flux, reaction; 28 epochs, batch 64), all trained with Adam (learning rate \(3\times10^{-3}\), orthogonality penalty 0.05 against frozen atoms, gradient-norm clip 2.0, seed \(100+s\)), totaling 7 stems, 21 blocks and 2.65M parameters. Latent states evolve over \(L=8\) forward-Euler residual updates of size \(1/L\), \(z^{(\ell+1)} = z^{(\ell)} + L^{-1}\sum_m \alpha_{\ell,m} A_m(z^{(\ell)}, h_{\mathrm{ctx}})\), and the final linear readout operates on the concatenated feature tensor \([z^{(L)}, x_{\mathrm{lin}}, x_{\mathrm{slog}}, 1]\).

\paragraph{Pre-enrollment adaptation and source removal.}
We evaluate transfer prior to target enrollment using the frozen Stage-6 checkpoint for Wave-Gauss (64 disjoint three-example support draws evaluated on a fixed 144-pair subset of the 240 held-out query pairs; 32 paired draws for the scratch comparison) and the frozen Stage-5 checkpoint for Acoustic Mazes (24 disjoint three-example support draws evaluated on all 48 held-out query pairs). With all stems and operator blocks frozen, we fit a step-shared mixing coefficient \(\alpha_m\) per frozen block alongside \((a_{\mathrm{in}}, b_{\mathrm{in}}, w_{\mathrm{stem}})\) for 55 Adam steps (learning rate 0.06, \(\ell_1\) penalty \(2\times10^{-3}\) on block and stem mixing weights, \(\ell_2\) penalty \(10^{-3}\) on these mixing weights and the channel-role matrices, gradient-norm clip 5.0) against leave-one-trajectory-out (LOTO) ridge regression error (\(\lambda=2\times10^{-2}\)) over the three support pairs. Using the lowest-LOTO adapter iterate, we select the final ridge penalty over \(\{10^{-4}, 3\times10^{-4}, 10^{-3}, 3\times10^{-3}, 10^{-2}, 3\times10^{-2}, 10^{-1}, 3\times10^{-1}, 1.0\}\) by LOTO error and refit the linear readout on all three support pairs.

For a source-removal intervention, we zero-mask \emph{both} the input stem and all operator blocks acquired at the specified source stage before repeating the identical adaptation on the exact same support draws and query fields. Source identities are used solely to construct the intervention mask and are never supplied as target physics labels during adaptation. Paired 95\% confidence intervals for error ratios use 5,000 bootstrap resamples over support draws (seed 251750). Reported ratios are ratios of draw-averaged errors, with numerator and denominator resampled jointly. On Wave-Gauss (64 draws), mean relative field error was 16.80\% intact, 45.27\% with Wave-Layer removed and 16.74\% with Darcy removed (random library 47.04\%). On Acoustic Mazes (24 draws), corresponding errors were 65.73\% intact, 98.24\% with Acoustic Inclusions removed and 64.38\% with Wave-Layer removed (random library 104.67\%). Joint stem-and-block masking leaves open how much of each gain originates in stem compatibility versus spatial operator weights.

\section{V3 training stages and evaluation}
\label{sec:s2}

Table~\ref{tab:s1} reports each field predictor at the stage that acquired it and an independently initialized target-only fit. For channel \(c\), \(\mathrm{VRMSE}_c = \sqrt{\langle(\widehat{y}_c - y_c)^2\rangle / \langle(y_c - \bar{y}_c)^2\rangle}\), with averages pooled over test samples and grid points; we report the mean of the channelwise ratios, each using its own reference standard deviation. The registered pair counts include multiple times from the same trajectory. For recursive tests we feed predictions back as inputs and evaluate step ten on up to 16 trajectories. The ocean target is a direct five-day anomaly forecast.

\begin{table}[!htbp]
  \centering
  \caption{\textbf{Registered pairs and channel-mean VRMSE at acquisition for the seven-stage V3 library.} OISST entries are VRMSE (n/a: no step-10 rollout for the direct five-day forecast); the main text reports mean relative \(L_2\) error for the same checkpoints (61.12\% continual, 59.07\% scratch).}
  \label{tab:s1}
  \small
  \begin{tabular}{@{}lrrrrrrr@{}}
    \toprule
    & \multicolumn{2}{c}{Field pairs} & \multicolumn{2}{c}{Step 1} & \multicolumn{3}{c}{Step 10}\\
    \cmidrule(lr){2-3}\cmidrule(lr){4-5}\cmidrule(l){6-8}
    Stage and target & Train & Test & Continual & Scratch & Continual & Scratch & Persistence\\
    \midrule
    1\quad Helmholtz       & 2,392 & 36    & 0.00030 & 0.00030 & 0.00191 & 0.00191 & 5.0920\\
    2\quad Shear flow      & 2,448 & 36    & 0.0423  & 0.0407  & 0.2564  & 0.4480  & 0.7795\\
    3\quad Rayleigh--B\'enard & 2,448 & 45 & 0.2320  & 0.2333  & 2.0106  & 1.8821  & 0.8291\\
    4\quad Active matter   & 884   & 27    & 0.4967  & 0.5242  & 1.4844  & 1.5506  & 1.3248\\
    5\quad Euler shocks    & 1,360 & 288   & 0.1655  & 0.1620  & 0.5340  & 0.5177  & 1.5190\\
    6\quad ERA5 atmosphere & 7,326 & 1,331 & 0.0916  & 0.0919  & 0.4277  & 0.4325  & 0.6705\\
    7\quad NOAA OISST      & 1,815 & 354   & 0.6205  & 0.5985  & n/a     & n/a     & n/a\\
    \bottomrule
  \end{tabular}
\end{table}

The seven-stage ocean checkpoint had 61.12\% mean relative \(L_2\) error versus 59.07\% from scratch and 49.76\% for its Stage-7 affine anchor (persistence 50.15\%). In a separate chronological fit (frozen six-stage library adapted on 2018--2021, early-stopped on 2022, anomaly climatology computed over 2018--2022), the 2023 errors were 49.41\% learned, 49.18\% scrambled, 49.02\% scratch and 49.77\% for the chronological affine anchor. Two unseen-family comparisons also delimit transfer. On two viscoelastic support trajectories the learned and scrambled libraries both gave 0.6560 one-step VRMSE (scratch 0.6559). On eight turbulent radiative layer trajectories the learned library gave 0.5079 versus 0.5104 scrambled and 0.4904 scratch. The simulation, atmosphere and ocean data derive from \emph{The Well}, ERA5 / WeatherBench~2 and NOAA OISST, respectively \cite{well,era5,wb,oisst}.

\clearpage
\section{Controlled V4 single-target accuracy at two horizons}
\label{sec:s3}

We train each model separately on each target for 1,600 optimizer steps with batches of 16 (single seed per model and target). All models use the same target data, channel standardization and fitted pixelwise affine anchor. The comparison holds update count fixed; parameter counts (listed in Table~\ref{tab:s2}), learning rates and objectives vary across architectures. The FNO2D, U-Net and rollout-trained V4 include a multi-step objective and a spectral loss, whereas V3 and one-step V4 fit only the one-step objective. We score one-step predictions at sampled test times, and recursive predictions from a common initial frame on at most 16 test trajectories.

\begin{table}[!htbp]
  \centering
  \caption{\textbf{Single-target VRMSE by model, family and forecast step.} Each row represents separately fitted parameters (single seed); bold marks the minimum within a column. Parameter counts vary slightly by input channel count (by less than 0.003M across targets).}
  \label{tab:s2}
  \small
  \begin{tabular}{@{}lrrrrrrr@{}}
    \toprule
    & & \multicolumn{2}{c}{Rayleigh--B\'enard} & \multicolumn{2}{c}{Euler shocks} & \multicolumn{2}{c}{ERA5}\\
    \cmidrule(lr){3-4}\cmidrule(lr){5-6}\cmidrule(l){7-8}
    Model & Params (M) & Step 1 & Step 10 & Step 1 & Step 10 & Step 1 & Step 10\\
    \midrule
    FNO2D               & 2.38 & 0.32592 & 0.89813 & 0.26349 & 0.70912 & 0.11517 & 0.44258\\
    U-Net               & 2.25 & 0.29660 & 0.85460 & 0.18818 & 0.46257 & 0.09299 & 0.38740\\
    V3                  & 2.11 & 0.23794 & 1.13519 & \textbf{0.17051} & 0.51285 & 0.09157 & 0.41597\\
    V4, one-step fit    & 1.51 & \textbf{0.23194} & \textbf{0.84617} & 0.17645 & 0.55436 & \textbf{0.07772} & 0.38647\\
    V4, rollout-trained & 1.51 & 0.25182 & 0.90338 & 0.18851 & \textbf{0.45908} & 0.08063 & \textbf{0.34738}\\
    \bottomrule
  \end{tabular}
\end{table}

As shown in Table~\ref{tab:s2}, the rollout objective improves tenth-step accuracy for Euler shocks and ERA5 relative to one-step V4, whereas for Rayleigh--B\'enard convection the one-step-trained V4 has lower error at both horizons.

\section{Transfer after the three-stage V4 acquisition}
\label{sec:s4}

We freeze the 12 blocks learned in sequence on Euler shocks, Rayleigh--B\'enard convection and ERA5, and give each target model the first two or four training trajectories from a held-out family. All three fits use the same target transitions, normalization, affine anchor and 350 updates. The learned and scrambled arms fit only a new input lift, spatial router and readout (about 0.136M trainable parameters). For the scrambled arm we independently permute the entries within every tensor of the stored blocks, preserving each tensor's marginal distribution. The target-only arm trains a new four-block V4 end to end (about 1.50M trainable parameters). These families were excluded from the three-stage V4 acquisition; their evaluation trajectories were also used in the V3 comparisons.

\begin{table}[!htbp]
  \centering
  \caption{\textbf{Held-out-family VRMSE after three-stage V4 acquisition.} Each support trajectory supplies multiple labeled times.}
  \label{tab:s3}
  \small
  \begin{tabular}{@{}lrrrrrrr@{}}
    \toprule
    & & \multicolumn{3}{c}{Step 1} & \multicolumn{3}{c}{Step 10}\\
    \cmidrule(lr){3-5}\cmidrule(l){6-8}
    Target & Traj. & Learned & Scrambled & Scratch & Learned & Scrambled & Scratch\\
    \midrule
    Planetary shallow water   & 2 & 0.056166 & 0.058518 & 0.040267 & 0.389502 & 0.401062 & 0.290062\\
    Planetary shallow water   & 4 & 0.052031 & 0.052874 & 0.034654 & 0.354172 & 0.331444 & 0.220946\\
    Turbulent radiative layer & 2 & 0.406766 & 0.405287 & 0.373448 & 0.938227 & 0.791963 & 0.503594\\
    Turbulent radiative layer & 4 & 0.405495 & 0.407253 & 0.368800 & 0.913450 & 0.838505 & 0.509530\\
    \bottomrule
  \end{tabular}
\end{table}

Table~\ref{tab:s3} shows that the learned blocks improve on their scrambled counterparts at both horizons for two planetary shallow-water trajectories; with four trajectories the tenth-step ordering reverses. On the turbulent radiative layer, scrambled blocks beat learned blocks at step 10 for both budgets, and end-to-end scratch has the lowest error in all four rows. These results bound what the frozen representation supplies under this adaptation budget.

\clearpage
\section{Inside the operator library}
\label{sec:s5}

\begin{figure}[!htbp]
  \centering
  \includegraphics[width=\linewidth,height=5.65in,keepaspectratio]{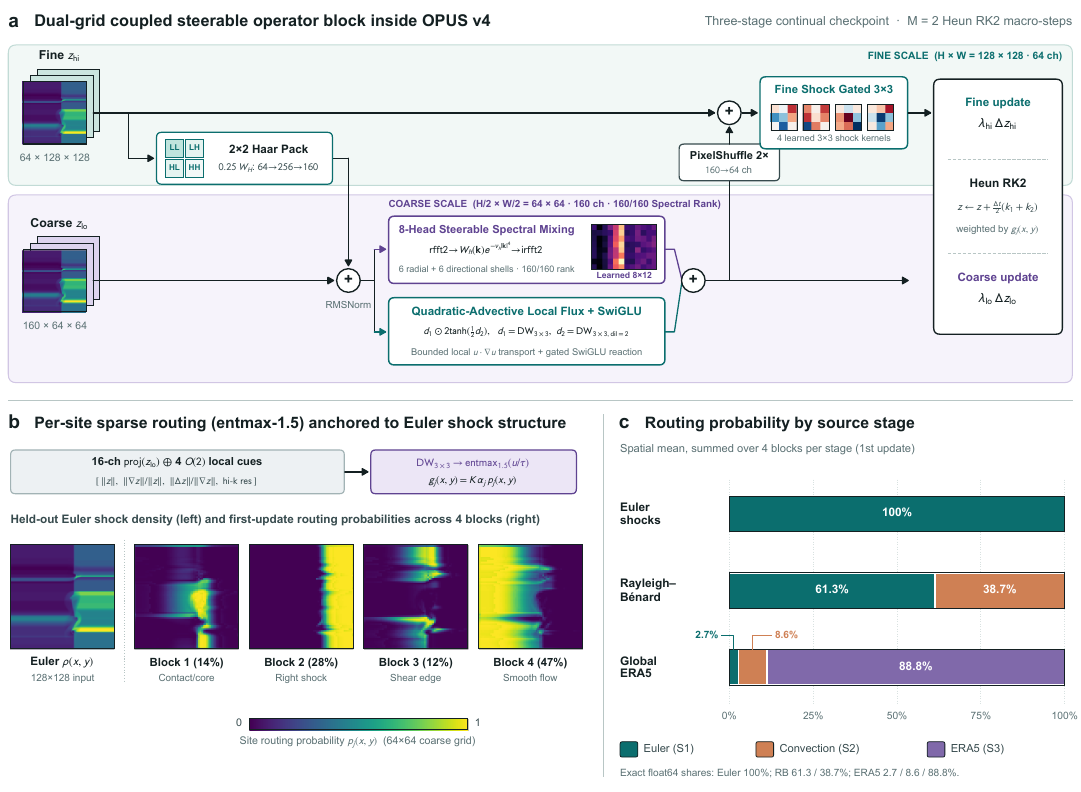}
  \caption{\textbf{Computation inside a V4 block and routing in the three-stage checkpoint.} \textbf{a}, The block couples 64 fine-grid channels to 160 coarse-grid channels; a \(2\times2\) Haar transform packs fine-grid features to \(4\times64\) channels and projects to 160, spectral and gated local branches update the coarse field, and \(2\times\) PixelShuffle returns features to the finer grid. \textbf{b}, First-update routing probabilities for the four eligible Euler blocks at the same \(64\times64\) coarse sites; the applied coefficient also includes each block's learned gain and the number of eligible blocks. \textbf{c}, Probabilities summed for each source's four blocks at every site and averaged across sites; each row is one held-out case.}
  \label{fig:operator-detail}
\end{figure}

In Figure~\ref{fig:operator-detail}, only Euler blocks are eligible for the Euler route, so their share is 100\% by construction. On the convection example the older Euler blocks receive 61.27\% of the probability and the convection blocks 38.73\%. On the ERA5 example the shares are 2.65\% Euler, 8.56\% convection and 88.78\% ERA5 (rounded). The effect of removing any block also depends on the field it produces and its learned gain; a V4 source-removal experiment, analogous to the V2 interventions in Figure~\ref{fig:interventions}, would test that effect directly.

V3 and V4 add a learned residual to an affine map fitted on training pairs,
\begin{equation}
  \widehat{y}_s(x) = \Gamma_s[x;1] + D_s\,R_s\!\left(\Phi_s(x;J_s)\right).
  \label{eq:residual}
\end{equation}
Here \(\Gamma_s\) is the pixelwise affine map, \(D_s\) scales the residual, and \(R_s\) reads out the target channels from the latent representation \(\Phi_s\). During enrollment V3 selects up to six earlier blocks and adds four, with learned gates acting by update step and latent channel. The three-stage V4 experiment admits the first eight earlier blocks at most and adds four per stage. Each of the two latent macro-steps is a Heun (RK2) step comprising two evaluations of the weighted update in Figure~\ref{fig:operator-detail}a. V2 uses a linear readout of latent, standardized and signed-log input features, with block coefficients varying by step during permanent enrollment and shared across steps during few-example adaptation.

\clearpage
\section{Field tasks in the evaluated checkpoints}
\label{sec:s6}

\begin{figure}[!htbp]
  \centering
  \includegraphics[width=\linewidth,height=5.65in,keepaspectratio]{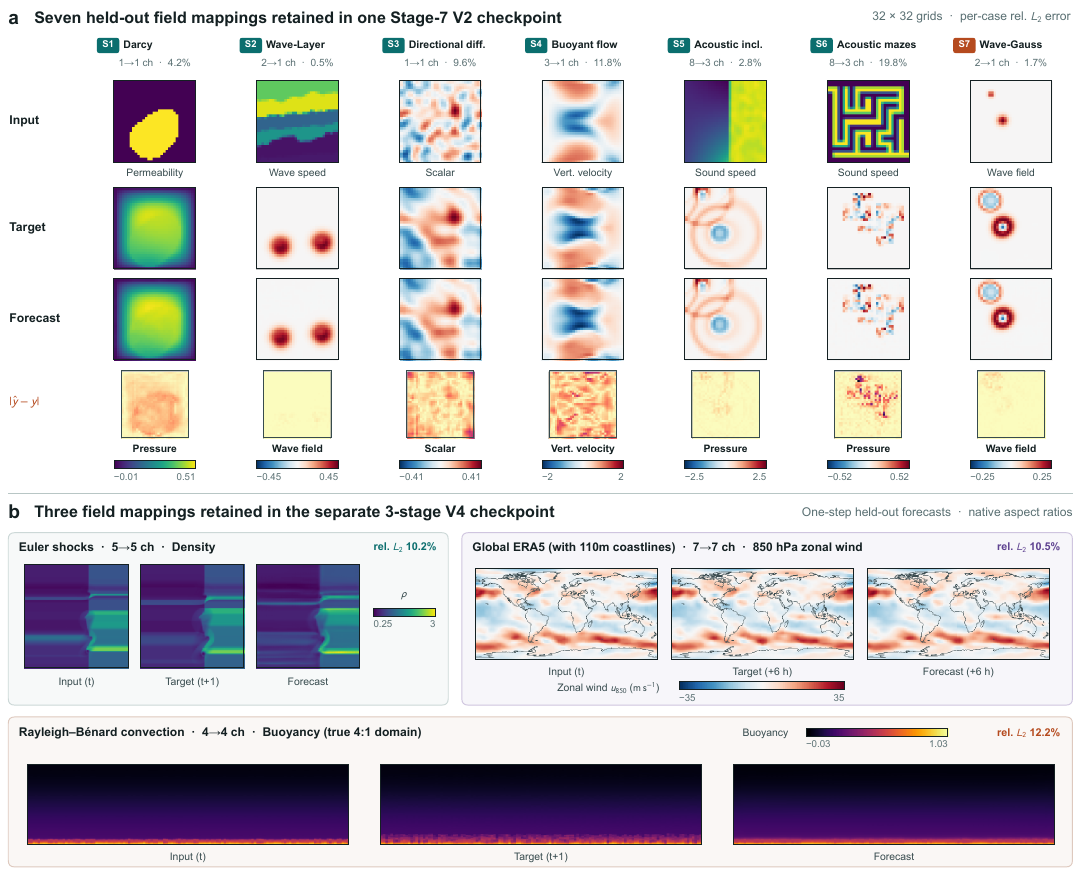}
  \caption{\textbf{Field mappings retained by two OPUS checkpoints.} \textbf{a}, Each column shows one selected input channel, one recorded target channel and the corresponding frozen prediction from the first case of a V2 evaluation panel. Arrows give the complete input and output channel counts. Each acoustic interface uses five current fields and three changes relative to an earlier measurement. The seven tasks share one Stage-7 checkpoint with seven input stems and 21 operator blocks. \textbf{b}, A separate three-stage V4 checkpoint predicts open-boundary Euler density, Rayleigh--B\'enard buoyancy and ERA5 zonal wind at 850~hPa. Each triplet shows one component of the input, its one-step reference and the archived prediction on the same trajectory. All targets and forecasts share the linear color scale shown for their task. Input ranges and trajectory identifiers are given in the figure data. These individual fields illustrate the interfaces; aggregate accuracy is reported separately.}
  \label{fig:task-atlas}
\end{figure}

Figure~\ref{fig:task-atlas} pairs reference and predicted fields from the V2 Stage-7 and three-stage V4 checkpoints. For the two acoustic examples, sound speed shows the material geometry while the displayed output is pressure. Both forecasts use the same V2 checkpoint with their respective stored settings; pressure error is 2.85\% for the inclusion case and 19.85\% for the maze case, compared with 7.02\% and 20.98\% across all 48 queries and three response channels in Table~\ref{tab:v2full}.

The seven V2 comparisons use \(32\times32\) registered cells. The V4 density and buoyancy fields use \(128\times128\) arrays, and the ERA5 wind uses a \(64\times128\) latitude--longitude grid upsampled from \(32\times64\) WeatherBench~2 data. The atmospheric target comes from reanalysis, while the other displayed targets come from archived simulations.

\end{document}